\documentclass{article}
\usepackage{booktabs, makecell, multirow}

\PassOptionsToPackage{numbers, compress}{natbib}
\usepackage[preprint]{neurips_2020}

\usepackage{caption}
\usepackage{subfigure}
\usepackage{xcolor}
\usepackage{epsfig}
\usepackage{graphicx}
\usepackage[utf8]{inputenc} % allow utf-8 input
\usepackage[T1]{fontenc}    % use 8-bit T1 fonts
\usepackage{hyperref}       % hyperlinks
\usepackage{url}            % simple URL typesetting
\usepackage{booktabs}       % professional-quality tables
\usepackage{amsfonts}       % blackboard math symbols
\usepackage{nicefrac}       % compact symbols for 1/2, etc.
\usepackage{microtype}      % microtypography
\usepackage{amsmath}
\usepackage{amssymb}
\usepackage{soul}
\usepackage{amsmath,bm}

\begin{document}
\title{Towards Generalized and Incremental Few-Shot Object Detection}

% The \author macro works with any number of authors. There are two commands
% used to separate the names and addresses of multiple authors: \And and \AND.
%
% Using \And between authors leaves it to LaTeX to determine where to break the
% lines. Using \AND forces a line break at that point. So, if LaTeX puts 3 of 4
% authors names on the first line, and the last on the second line, try using
% \AND instead of \And before the third author name.

\author{
	Yiting Li\textsuperscript{\rm 1},\quad
	Haiyue Zhu\textsuperscript{\rm 2}\thanks{Corresponding author},\quad
	Jun Ma\textsuperscript{\rm 1},
\AND
	Chek Sing Teo\textsuperscript{\rm 2},\quad %\IEEEmembership{Student Member,~IEEE,}
	Cheng Xiang\textsuperscript{\rm 1},\quad
	Prahlad Vadakkepat\textsuperscript{\rm 1},\quad and
	Tong Heng Lee\textsuperscript{\rm 1}\\
 \textsuperscript{\rm 1} National University of Singapore\\
 \textsuperscript{\rm 2} SIMTech, Agency for Science, Technology and Research\\
}

\maketitle
\begin{abstract}
Real-world object detection is highly desired to be equipped with the learning expandability that can enlarge its detection classes incrementally. Moreover, such learning from only few annotated training samples further adds the flexibility for the object detector, which is highly expected in many applications such as autonomous driving, robotics, etc. However, such sequential learning scenario with few-shot training samples generally causes catastrophic forgetting and dramatic overfitting. In this paper, to address the above incremental few-shot learning issues, a novel Incremental Few-Shot Object Detection (iFSOD) method is proposed to enable the effective continual learning from few-shot samples. Specifically, a Double-Branch Framework (DBF) is proposed to decouple the feature representation of base and novel (few-shot) class, which facilitates both the old-knowledge retention and new-class adaption simultaneously. Furthermore, a progressive model updating rule is carried out to preserve the long-term memory on old classes effectively when adapt to sequential new classes.
Moreover, an inter-task class separation loss is proposed to extend the decision region of new-coming classes for better feature discrimination. We conduct experiments on both Pascal VOC and MS-COCO, which demonstrate that our method can effectively solve the problem of incremental few-shot detection and significantly improve the detection accuracy on both base and novel classes.
\end{abstract}

\section{Introduction}
Modern deep-learning based object detection approaches~\cite{NIPS2015_FasterRCNN,Bochkovskiy_2020_Yolo4,cascade_rcnn} have achieved remarkable results on benchmarks such as Pascal VOC and MS-COCO datasets. This, however, usually requires a large amount of labeled data, which can be labor-intensive and time-consuming to acquire. On the contrary, humans can recognize and locate new objects after observing only one or a few instances. Such ability to learn from few examples is desirable for many applications in a low-data regime where the labeled data is rare and hard to collect. The challenge of detecting novel classes from few samples is usually referred to as Few-Shot Object Detection (FSOD), which is usually framed as either metric learning or meta-learning problems in previous studies. For instance, TFA~\cite{Wang2020}
is one of the most representative approaches that adopting metric learning into few-shot detection.  Meta-RCNN~\cite{Yan2019MetaRCNN} propose a meta-learning attention mechanism to emphasize correlated feature channels for better classification. Despite their successes, when extending to novel classes, they require to store and replay a memory buffer for old classes, which often results in heavily model retraining and inevitably hinders their real-world applications.

In this paper, we study a more challenging problem sets, i.e., Class-Incremental Few-Shot Object Detection (CI-FSOD)~\cite{Once}. Unlike the conventional FSOD methods, CI-FSOD requires not only to reduce the hardware resource requirement but also to retain the comparable detection performance. Hence, a desirable solution should contain three important aspects. First, the model can learn efficiently only from a continual data stream without catastrophic forgetting. For example, when the model adapts to new tasks, it cannot be fine-tuned with previous task data to achieve fast adaption and storage saving. Second, the model should generalize well even when the volume of training samples is small for novel classes. Third, previous knowledge gained from the large-scale base training set should be well preserved. It is noted that many previous approaches only focus on promoting feature representation for novel classes but fail to achieve good knowledge retention on the base classes.

Catastrophic forgetting affects the training of deep models, limiting their abilities to learn multiple tasks sequentially. To tackle catastrophic forgetting for the large-scale base classes to maintain the overall performance, TFA only fine-tunes the last two $fc$ layers on a small balanced training set containing both base and novel classes, while the feature extractor is kept to be fixed. This could be viewed as a practical approach to preserve the previously learned feature distribution for base classes. However, it is unreasonable that the pre-trained RoI feature extractor, which contains much semantic information of base classes, could be used directly to represent unseen novel classes without further fine-tuning. In contrast, when we fine-tune TFA by unfreezing RoI feature extractor, it is surprising to see that the AP performance on novel classes could be significantly improved from $9.9$ to $12.3$ points. Such findings indicate that fine-tuning deeper layers of CNN are crucial for learning unseen concepts since it can obtain better features. However, the worse results on base class(decreasing from $28.8$ to $26.5$ points) reveal that fine-tuning base classes with limited data may inevitably lead to catastrophic forgetting of the originally generalizable representation~\cite{BBN}. To better preserve base classes' performance and incorporate our method with non-incremental settings, we propose to decouple the representation learning of base and novel classes into two independent branches. In particular, we propose a novel Double-Branch Framework to take care of both base-class retention and novel-class adaption simultaneously.

Upon overcoming knowledge retention on base classes, we are still facing catastrophic forgetting on those sequentially encountered novel classes. In this work, we find that such issues can be attributed to two reasons. First,for models learn in a few-shot sequential manner, overfitting due to data scarcity may further exacerbate catastrophic forgetting. According to the well-established plasticity-stability theory ~\cite{understanding_IL}, model not only relies on plasticity to adapt quickly to novel tasks but also requires stability to prevent catastrophic forgetting for old tasks. From a loss landscape perspective, the flatness of the local minima found for each task plays an essential role in the overall degree of forgetting. Especially, wide and flat local minima is desired for each incremental task, so as to guarantee that minima found for different tasks could overlap with each other to counter catastrophic forgetting, ~\cite{understanding_IL}. However, under few-shot condition, optimizing neural networks on small training sets is almost same as adopting the large-batch training regime, which tends to converge to narrow and sharp local minima of the training and testing functions—and as is well known, sharp minima lead to poorer generalization~\cite{CPR}. As a result, networks trained from  few-shot tasks tend to be less stable and forget previously seen tasks more severely. A common approach Incremental Moment Matching (mean-IMM)~\cite{IMM} proposes to prevent catastrophic forgetting by averaging model weights of two sequential tasks, by implicitly assuming that the obtained optimum at each incremental step is flat and overlaps with each other. However, we argue that such assumption can only hold for data-sufficient scenarios but not for few-shot. Given abundant training samples, the overall incremental-learning process is unlike to suffer from the large-batch training dilemma. Even without any additional regularization, the obtained local minima at each step can still be flat and wide, thus model with flatter minima is more generalizable to unseen classes as it's always more likely to overlap with the other optimums, which is not realistic for  few-shot scenarios.

Second, in an incremental learning manner, a model can only access the data of the current task since no data of previous tasks can be stored and replayed. As a result, inter-class separation can only be achieved inside the current task, while the discrimination between inter-task classes is missed. For example, if the first task is to discriminate truck vs car and the second task is to classify bus vs airplane, no discriminative feature learned specifically to distinguish truck from the bus.

In this work, we consider tackling the issues above from two perspectives. 1) Optimization and loss landspace: We propose a novel Stable Moment Matching (SMM) algorithm for CI-FSOD setting. In particular, the pretrained base-set weights can serve as good parameter initialization because of its flatness, our goal is to pass its flatness among the sequential-encountered few-shot tasks, through restricting parameter updating process within the same local flat region. In particular, we reinforce the training stability by exerting a stronger resistance force towards model updating, by restricting overall parameter displacement to be upper-bonded. To this end, the searching space of parameter updating to a small local region around optimum of the previous task, thus high intersection of low loss surface between different tasks can be achieved. 2) Inter-task class separation: We propose to store RoI features of previous tasks rather than the original images for representing old classes. A margin-based regularization loss is proposed to optimize margins for the frequently misclassified old classes.
%(3)	To better preserve performance on base classes and also incorporate our method with non-incremental settings, we propose to decouple the representation learning of base and novel classes for the CI-FSOD problem. In particular, we propose a novel Double-Branch Framework to take care of both base knowledge retention and novel knowledge adaption simultaneously.

\section{Related Works}
%\subsection{Incremental Learning}
%Recent research that focuses on how to remedy catastrophic forgetting can be roughly grouped from two aspects: data replay and model regularization. The first group of methods usually store a memory of the learned knowledge from historical tasks. For example, iCaRL builds a class-representative memory with exemplars that are close to the category center. BiC propose to learn a bias correction module from an unbiased validation set to eliminate prediction bias between old and new tasks. Regularization-based methods pay attention to stabilize parameters updating during the learning process. For example, EWC exerts a constrain on the alterations of model parameters according to the importance of weights in previous learned tasks given by fisher information. Incremental Moment Matching (IMM) restricts the parameters updating for the current task to be close to the local minima of the previous task, an extra L2 regularization term is added to preserve linear connectivity between multiple tasks. However, to achieve \hl{long-term memory mainly}, they require storing either Fisher matrix or model weights for each task independently, which is practically infeasible if there are many tasks and the network has millions of parameters.

\textbf{Incremental Few-Shot Learning}
Incremental Few-shot learning is attracting increasing attention due to its realistic applications~\cite{Chelsea_2017_MAML,Snell_2017_Prototypical}. However, most of the existing methods are proposed to address the single image classification problem, hence not readily applicable to object detection. Our method falls within the context of the regularization-based learning approach~\cite{IDQ,IMM,EWC}. For example, when learning incrementally, ILDVQ~\cite{IDQ} preserves feature representation of the network on older tasks by using a less-forgetting loss, where response targets are computed using data from the current task. As a result, ILDVQ does not require the storage of older training data. However, this strategy may be inefficient if the data for the new task is too few or just belongs to a distribution different from those of prior tasks.  In contrast, we address the incremental few-shot learning problem from a new perspective of parameter space. The proposed Stable Moment Matching algorithm strengthens the stability of few-shot adaptation and is more robust to data scarcity.

\textbf{Few-Shot Object Detection}
The most recent few-shot detection approaches are adapted from the few-shot learning paradigm. \cite{Hao2018LSTD} proposes a distillation-based approach with background depression regularization to eliminate the redundant amount of distracting backgrounds. A meta-learning attention generation network is proposed in \cite{Kang_2019_YOLOLS} to emphasize the category-relevant information by reweighting top-layer feature maps with class-specific channel-wise attention vectors. Sharing the same insight, Meta-RCNN~\cite{Yan2019MetaRCNN} applies the generated attention to each region proposal instead of the top-layer feature map. TFA~\cite{Wang2020} replaces the original classification head of Faster-RCNN with a cosine classifier to stabilize the adaptation procedure. %However, existing FSOD methods mainly consider the non-incremental learning setting. When new classes are to be added, they rely on the base training dataset during fine-tuning, which dramatically increases memory space and computation cost, and restricts their scalability into realistic applications such as IoT and robotics.

\section{Methods}

\begin{figure}
	\centering
	\includegraphics[width=0.8\columnwidth]{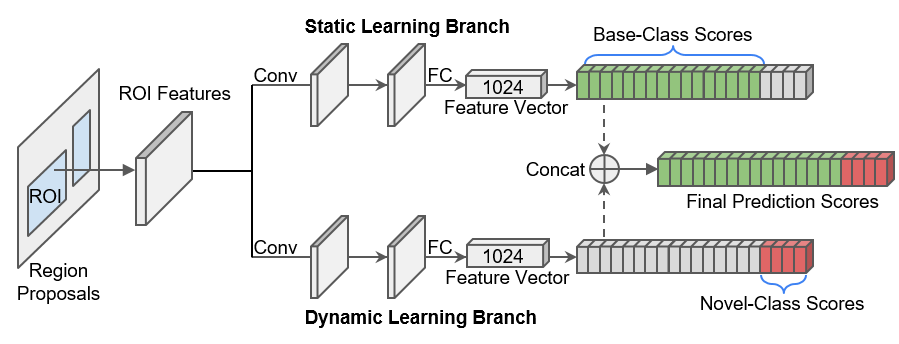}
	\caption{Illustration of the proposed Double Branch Framework. 1) The static learning branch only updates its new-added weights on the box classifier in the current task, responsible for preserving the original base-class feature distributions. 2) The dynamic learning branch jointly updates its RoI feature extractor, box regressor, and the new-added box classifier weights, designed to model the new-coming classes better. Finally, corresponding parts of the prediction confidence vectors from two branches are sliced and concatenated for the overall prediction scores.}
	\label{fig:DBF}
\end{figure}
\subsection{Problem Formulation}

The CI-FSOD problem is normally formulated as a two-phase learning task. In the first representation learning phase ($n=0$), a detection model $F_{0}(\cdot|\bm{W_{0}})$ is pretrained on a large set of base classes where $\bm{W_0}$ denotes the learned weight parameters. In the second incremental learning phase ($n>0$), given the sequential new few-shot task $\mathcal{T}_n$, the model updating $F_{n}(\cdot|\bm{W_{n}})$ is performed over multiple stages from $\mathcal{T}_{1}$ to $\mathcal{T}_{n}$. During the $n$-th learning step, the previous category space $\mathcal{Y}_{n-1}$ is expanded with new classes $\mathcal{C}_n$ in $\mathcal{T}_{n}$, so that $\mathcal{Y}_n = \mathcal{Y}_{n-1}\cup \mathcal{C}_n$, where $\mathcal{Y}_{n-1}\cap \mathcal{C}_n=\emptyset$ is assumed for simplicity. The objective of CI-FSOD is to effectively learn a $F_{n}(\cdot|\bm{W_{n}})$ which is capable to detect all the classes in $\mathcal{Y}_{n}$ with high accuracy. In this work, we use ``base'' and ``novel'' classes to differentiate the classes in representation learning phase ($n=0$, not few-shot manner) and incremental learning phase ($n>0$, few-shot manner). Moreover, we use ``new'' and ``old'' classes to denotes the new coming classes in $\mathcal{T}_{n}$ and previous few-shot classes from $\mathcal{T}_{1\,\text{to}\,n-1}$, noted that both of them are belong to novel classes.

\subsection{Overall Framework}

In this work, a novel DBF is proposed to address the balance problem between base-class knowledge retention and novel-class knowledge adaptation for iFSOD. The schematics of DBF is shown as in Fig.~\ref{fig:DBF}, where the base detector here we use Faster-RCNN as an illustration example. It is also noted the proposed DBF framework is also compatible to other base detectors such as YOLO, etc. The speciality of Fig.~\ref{fig:DBF} exists on its ROI feature extractor and sibling head (category classification and bbox regression),
where DBF is implemented with two parallel branches, termed as the ``Static Learning Branch” and ``Dynamic Learning Branch”, respectively.

The static learning branch adopts a similar feature consolidation strategy as TFA~\cite{Wang2020}. When a new task $\mathcal{T}_n$ comes, we freeze its backbone, RPN, ROI feature extractor, old-class classification weights and box regressor, while only expands the classifier weight matrix with new-class weights and then fine-tunes it with cosine similarity. Thus, the original feature representation is preserved to best represent base classes. In contrast, we unfreeze the ROI feature extractor for dynamic learning branch to better adapt to novel classes, together with a proposed SMM updating rule (Section III.B), an inter-task class regularization (Section III.C) and a semi-supervised pseudo labeling approach (Section III.D) to better prevent catastrophic forgetting when learning with long-term sequential tasks. After that, the final predicted proposal classification output is concatenated from the two branches as illustrated in Fig.~\ref{fig:DBF}, where the base-class scores are from the static learning branch and the novel-class scores are from the dynamic learning branch. Such a combination provides a better unbiased prediction of the overall learned classes in iFSOD setting.

\subsection{Stable Moment Matching (SMM)}
Currently, existing regularization-based approaches such as EWC ~\cite{EWC} and IMM ~\cite{IMM} alleviate the catastrophic forgetting effect by parameter consolidation. Although these methods have shown promising results when training data is sufficient, performance in few-shot scenario is shortlived. Moreover, they cannot achieve long-term memory unless previous fisher matrix or model weights are preserved, i.e, good performance can only be achieved for the first a few incremental steps, but performance rapidly decays during the following steps. Without loss of generality, we analyze the failure cause of IMM for the case of multi-step few-shot detection.

\begin{figure}
	\centering
	\includegraphics[width=0.9\columnwidth]{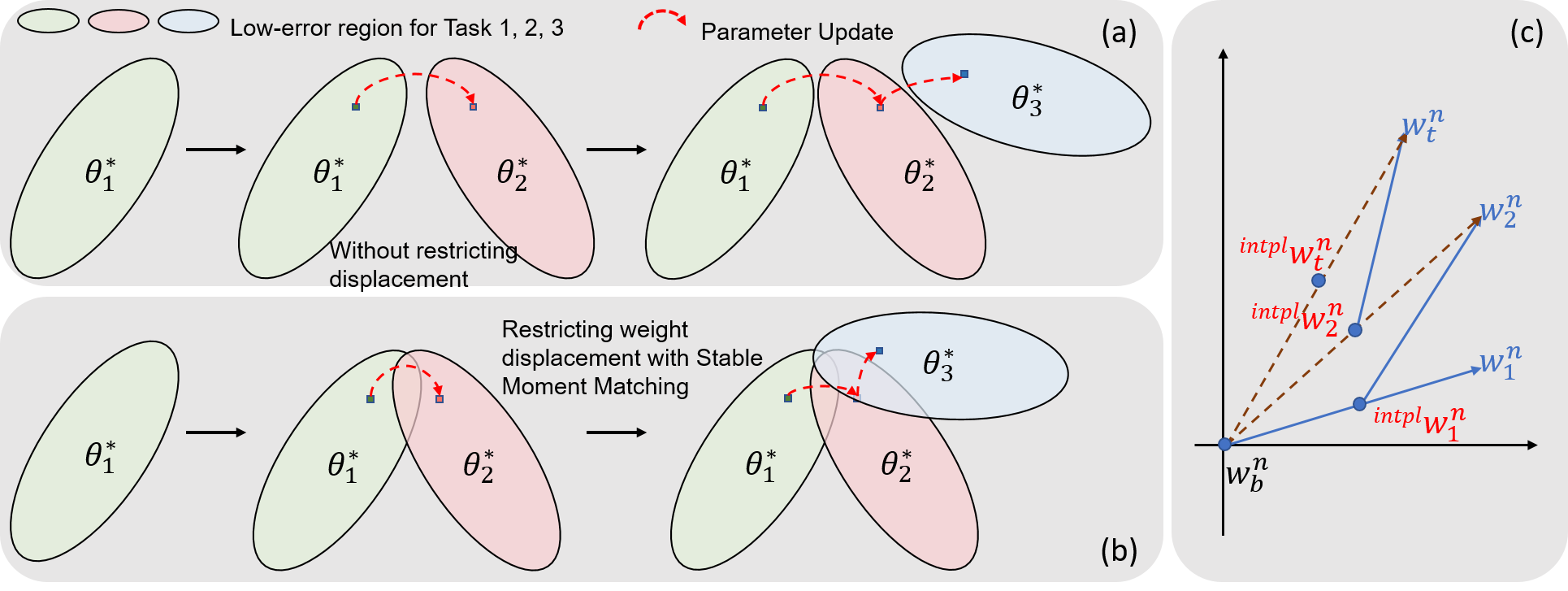}
	\caption{(a) In data-rare cases, the low-error surface around each local minima is sharp and narrow. Thus, the likelihood for these low-error surface overlapping becomes very small when the learning continues, dramatically increasing the difficulty of finding optimal solutions that perform well on different tasks. (b) Our intuition is that by restricting the magnitude of parameter displacement for each task, the obtained local minima can stay closer, allowing more ellipsoids to overlap. (c) Linear interpolation with the knowledge base model is performed in each iteration with an adaptive ratio in SMM. }
	\label{fig:WeightDisplacement}
\end{figure}

IMM assumes there exist a linear connector without high loss barriers between the local minima of two sequential tasks, so we can expect their averaging point to perform well for both tasks. However, we argue that such linear connectivity can only be guaranteed when the loss surface around local minima is flat~\cite{understanding_IL,linearmultiple,linearconnectivity}, which allows the low-error ellipsoid around each optimum to be wide enough to overlap with each other. However, when fine-tuning with extremely small data, a model is often suffering from large-batch training dilemma~\cite{keskar2016large}, which results in sharp and narrow minima. Thus the chance of existing loss-smooth linear connector between different tasks quickly becomes very small as the learning continues, illustrated as in Figure~\ref{fig:WeightDisplacement} (a). Model is often overfitted and over-plastic (forgetting quickly).

Given the pretrained detector, it’s supposed to have a much flatter local minima since it’s trained from abundant training data. Thus, our intuition is that, to ensure the flatness of local minima found for future few-shot tasks, the overall optimization trajectory should be restricted strictly to stay within the local-flat region of the pretrained base-set minima. In that way, linear connectivity among different tasks can be guaranteed, and the ellipsoids of different local minima can always overlap with each other even when learning continues.

\begin{figure}
	\centering
	\includegraphics[width=1\columnwidth]{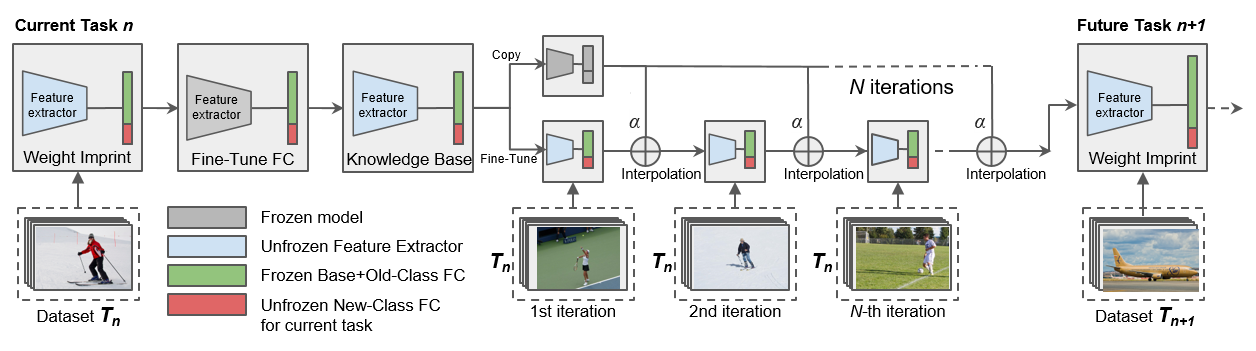}
	\caption{Illustration of the proposed Stable Moment Matching (SMM). Knowledge base is obtained by only fine-tuning the new-added weights in the box classifier. SMM simply interpolates the parameters of two models during each iteration. Once finishing the current task, we use the last interpolated model as the new knowledge base for next task. }
	\label{fig:SMM}
\end{figure}
Assume the local minima obtained through SGD from two continual tasks $\mathcal{T}_{n-1}$ and $\mathcal{T}_{n}$ are $\bm{w}^{n-1}$ and $\bm{w}^{n}$, respectively, where $\bm{w}$ represents model parameters contained in RoI feature extractor, box classifier, and box regressor in CI-FSOD. IMM simply adopts parameter interpolation as,
\begin{equation}
\label{eq:IMMinterpolation}
\begin{aligned}
\bm{w}^ \ast = \alpha\bm{w}^{n-1}+(1-\alpha)\bm{w}^{n},
\end{aligned}
\end{equation}
\normalsize
where $\alpha$ is the interpolation ratio. If $\alpha=0.5$, $\bm{w}^ \ast$ becomes the mean of $\bm{w}^{n-1}$ and $\bm{w}^{n}$. However, IMM ignores the fact that the existence of the linear connectivity between two local minima is highly dependent on the small overall displacement between $\bm{w}^{n-1}$ and $\bm{w}^{n}$, defined as $\Delta \bm{w_{n-1}^{n}} = ||\bm{w}^{n-1} - \bm{w}^n||^2$.  In this work, we propose an optimized learning regime by exerting a stronger resistance force on the overall displacement between the sequentially learned local minima. In particular, the proposed Stable Moment Matching (SMM) improves IMM by continually updating model weights through parameter interpolation for each iteration with an adaptive interpolation ratio. Specifically, the representation learning of each task are decoupled into two phases: classifier learning phase $Phase_{fc}$ and representation learning phase $Phase_{ex}$.

In $Phase_{fc}$, assuming the model weights obtained from the previous task $\mathcal{T}_{n-1}$ is $\bm{w}^{n-1}$, when a new task $\mathcal{T}_{n}$ comes, we adopt the commonly used weight imprinting strategy ~\cite{Hang2018} to initialize the classification weights for the new classes contained in $\mathcal{T}_{n}$. Then, we freeze the backbone, RPN and RoI feature extractor, and only fine-tune the new imprinted weights in the $fc$ classifier as well as the box regressor. The obtained model $\bm{w}^{n}_{b}$, denoted as knowledge base, preserves previous task knowledge to the best content thus can be used as a base model for $Phase_{ex}$. In $Phase_{ex}$, we unfreeze the RoI feature extractor and use the knowledge base $\bm{w}^{n}_{b}$ for iterative linear interpolation. Once finish the $t$-th gradient descent update, the linear interpolation is conducted between the knowledge base $\bm{w}^{n}_{b}$ and the newly updated weights $\bm{w}^{n}_t$ which obtained from the latest $t$-th single gradient descent,
\begin{equation}
\label{eq:interpolation}
\begin{aligned}
^{intpl}\bm{w}^{n}_{t} = \bm{w}^{n}_{b}\cdot \alpha +\bm{w}^{n}_{t}\cdot(1 -\alpha),
\end{aligned}
\end{equation}
\normalsize
where $\alpha$ is the interpolation ratio. Iteratively, the obtained interpolated weights $^{intpl}\bm{w}^{n}_{t}$ is then used as the starting point for the next $t+1$ iteration, where the process is visualized in Figure~\ref{fig:WeightDisplacement} (c). Specifically, given a mini-batch of images and a learning rate $\eta$ , the model is updated according to the gradient descent for one iteration,
\begin{equation}
\label{eq:update}
\begin{aligned}
\bm{w}^{n}_{t+1} =\ ^{intpl}\bm{w}^{n}_{t} -  \eta  \cdot \nabla \bm{L},
\end{aligned}
\end{equation}
\normalsize
where $\bm{L}$ represents the overall loss function. During the training of each task, the knowledge base model is fixed to be the same until converging. Only when a new task $\mathcal{T}_{n+1}$ comes, the knowledge base model $\bm{w}^{n+1}_{b}$ will be updated accordingly as mentioned in $Phase_{fc}$. The overall flowchart of the proposed SMM is illustrated in Figure~\ref{fig:SMM}.

From the standpoint of parameter space, an observation can be made for the total displacement $\Delta \bm{w}_{n-1}^{n}$ of model parameters for a $n$-th new task. Assuming the learning step length $|\eta  \cdot\nabla \bm{L}|$ is bounded by some constant $|\nabla \bm{L}|_{max}$, the upper bound of $\Delta \bm{w}_{n-1}^{n}$ for SMM with $N$ iterations of fine-tuning can be derived using recursion with \eqref{eq:interpolation} and \eqref{eq:update} as,
\begin{equation}
\begin{aligned}
 \big|\Delta \bm{w}_{n-1}^{n}\big|_{max}^{SMM} = \sum_{t=1}^{N}  (1-\alpha) ^t |\bm{L}|_{max}=\frac{\big(1-(1-\alpha)^{N}\big)|\bm{L}|_{max}}{\alpha}\leq \frac{|\bm{L}|_{max}}{\alpha}, for\;  0<\alpha<1.
\end{aligned}
\end{equation}
\normalsize
which indicates that the displacement of our approach is upper-bounded by$\frac{|\bm{L}|_{max}}{\alpha}$. By Similarly, according to the mean-IMM approach, its corresponding upper bound of $\Delta \bm{w}_{n-1}^{n}$  with $N$ iterations of fine-tuning  can be derived as,
\begin{equation}
\begin{aligned}
 \big|\Delta \bm{w}_{n-1}^{n}\big|_{max}^{mean-IMM} = (1-\alpha)\sum_{t=1}^{N} |\bm{L}|_{max}=(1-\alpha)N|\bm{L}|_{max}.
\end{aligned}
\end{equation}
\normalsize
Obviously, it can be easily seen that when $N$ increases, $ \big|\Delta \bm{w}_{n-1}^{n}\big|_{max}^{mean-IMM}> \big|\Delta \bm{w}_{n-1}^{n}\big|_{max}^{SMM}$ is easy to hold as the upper-bound of our method is independent of the iteration number $N$, which indicates the proposed SMM can provide smaller and controllable parameter displacement than mean-IMM for existence of the linear connectivity, illustrated as in Figure~\ref{fig:WeightDisplacement} (b). Moreover, a L2 regularization term ~\cite{IMM} for further promoting the linear connectivity between the current model $^{intpl}\bm{w}^{n}_{t}$ and its corresponding knowledge base $\bm{w}^{n}_{b}$ can be defined as $L_{reg} =||^{intpl}\bm{w}^{n}_{t} -  \bm{w}^{n}_{b}||^2$.

With a constant learning step length, the interpolation ratio $\alpha$ will determine the distance of the overall trajectory. A smaller $\alpha$ may increase the upper bound of the overall displacement from the beginning of training to bring more plasticity. However, using a small interpolation rate means applying a larger update to the model weights for a sequential learning problem. Therefore, the resultant model may learn fast but also forget quickly. Hence, we propose an adaptive interpolation ratio that starts with a small interpolation rate at the beginning of training. Then, we slightly increase the interpolation rate for each subsequent task to stabilize the optimization trajectory and prevent forgetting. Precisely, the adaptive interpolation ratio $\alpha$ for the $n$-th task in $q$-th training epoch is calculated as,
\begin{equation}
\begin{aligned}
\alpha(n, q) = (r_{step}\cdot n +r_{base})\cdot q \Big{/} N_{epoch},
\end{aligned}
\end{equation}
where $r_{base}$ and $r_{step}$ are the base rate and step increasing rate. $N_{epoch}$ represents the total number of training epochs for each task. Note that the current epoch number $q$ will be reset to $0$ at the beginning of each task.

\normalsize
\begin{figure}
	\centering
	\includegraphics[width=0.9\columnwidth]{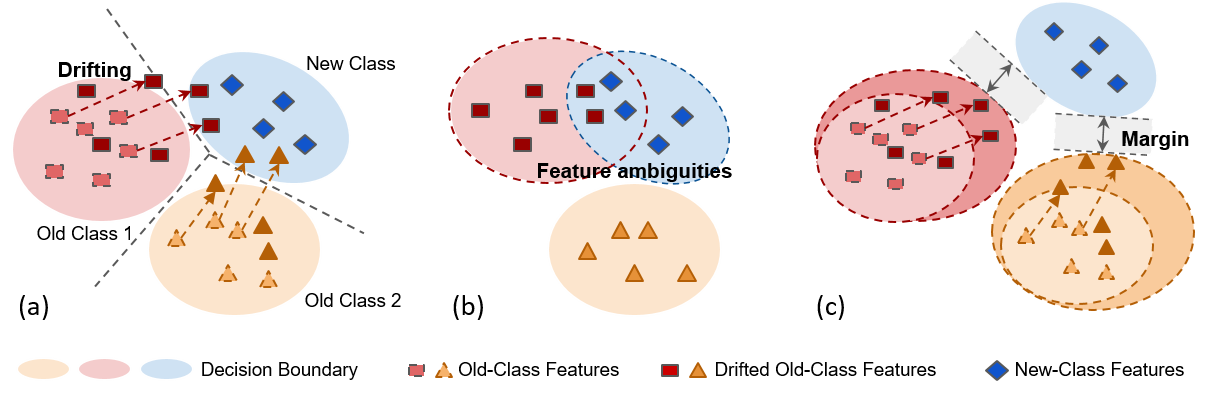}
	\caption{(a) Category confusion caused by feature drift. Samples of old tasks are well separated from new tasks initially. However, when a model is biased towards new-task classes during the fine-tuning, features of old classes are drifted and may be mixed up with those new classes. (b) Category confusion is caused by similar visual classes, such as different categories of dogs. (c) The proposed inter-task separation loss assigns more decision margins for the drifted old classes and promotes a more compact representation for current-task new classes.}
	\label{fig:featuredrift}
\end{figure}

\subsection{Inter-Task Separation Loss}
During the long-sequential learning process, training samples of previous tasks cannot be stored due to computation cost and privacy safety. As a result, inter-class separation is only optimized inside the current task, while the separation among classes from different tasks is neglected. This could further cause two problems: 1) Old classes of previous tasks may suffer from feature drift as the newly adapted feature extractor is biased towards new classes from the current task. Thus the previously learned classification weights for old classes are no longer compatible to the current embedding space~\cite{semantic_drift,IDQ}. 2) feature regions of novel classes in a new task could overlap with those old classes if similar discriminative features are shared, leading to more ambiguity among classes in the feature space. Such issues are conceptually illustrated in Figure~\ref{fig:featuredrift} (a) and (b), respectively.

Eliminating such inter-task class ambiguities is challenging as it requires to recall previous images for all old classes. To address the issues above, we propose a more feasible approach that collects representative feature embeddings from the last classification layer rather than storing raw images, so as to save the knowledge of the previous incremental steps and make the model to achieve better discrimination among all classes. Moreover, our method does not need previous classes’ images when adapting to new classes, which is more memory-efficient and privacy-secure. Although feature space generally suffers from drifting along multiple incremental steps, such drift can be effectively restricted to minimum by our proposed SMM regularization. Therefore, the stored historical embedding can be used as a practical approximation to represent old-class feature distributions.

To better promote inter-class discrimination, a margin-based separation loss is proposed to push the decision regions of novel classes further away from the features of old-classes, which not only encourages a more compact representation but also alleviate category confusion. This concept is illustrated in Figure~\ref{fig:featuredrift} (c). Specifically, the foreground RoI feature embeddings are sampled during each historical incremental task ($\mathcal{T}_{i=1\,\text{to}\,n-1}$) and stored as $\bm{X}_{old}=\{f(\bm{x})\}$, where $f(\bm{x})$ denotes the extracted feature for the corresponding RoI proposals. Given a new-task few-shot dataset $\mathcal{T}_n$, we randomly sample a fixed number of their foreground RoI features $\bm{X}_{new}=\{f(\bm{x})\}$ through each image. In the meantime, an equal number of historical feature embeddings are sampled from $\bm{X}_{old}$ for each old class. With the cosine similarity,if a RoI feature $x$ belongs to the new classes, we use the normalized weights of its ground-truth class weight vector $\theta_{new}^{+}$ as a positive template, and the negative template $\theta_{old}^{-}$ is selected from the old classes' weight vector that yields the highest similarity to $f(\bm{x})$, while the same arrangement is applied to those old-class feature embeddings. Mathematically, the proposed inter-task separation loss can be formulated as
\begin{equation}
\begin{aligned}
 L_{mgn}= & \sum_{\bm{x}_n \in {New}} \max \big(m- \kappa \cdot \theta_{new}^{+} \cdot f(\bm{x}_n)+ \kappa \cdot \theta_{old}^{-} \cdot f(\bm{x}_n),\ 0 \big)\\&
 +\sum_{\bm{x}_{o} \in {Old}} \max \big(m- \kappa \cdot \theta_{old}^{+} \cdot f(\bm{x}_{o})+ \kappa \cdot \theta_{new}^{-} \cdot f(\bm{x}_{o}),\ 0 \big), \end{aligned}
\end{equation}
\normalsize
where $m$ is a margin, $\kappa$ is a  scale parameter used to ensure the convergence of training~\cite{cosface}, and $\theta_{new}^{+}$ and $\theta_{old}^{+}$ are the positive templates for new-class and old-class proposals, respectively, using the weight vectors for the ground-truth classes. $\theta_{old}^{-}$ and $\theta_{new}^{-}$ are the negative templates for new-class and old-class proposals, respectively, using the highest-similarity weight vectors from the other groups, i.e., old and new classes.

\section{Training Strategy}
After finishing the pre-training, we freeze the backbone and RPN as their weights are considered to be category-agnostic. In each step of few-shot adaption, when a training image is feeded, the image passes through the backbone feature extractor and RPN to get ROIs. After the ROI-align layer, the obtained ROI features are passed into the following double branches for classification weights imprinting and model training. Since the goal of static learning batch is to best preserve base-class performance without forgetting, only the newly imprinted weights for the current incremental step are fine-tuned, while the backbone, RPN, ROI feature extractor, classification weights belonging to old classes as well as the regression weights are kept fixed. For the dynamic learning branch, to better adapt pre-trained features to represent novel classes, we further unfreeze the ROI feature extractor and update its weights with the proposed SMM rule. The overall learning objective for the dynamic learning branch contains three components, $L_{total}= L_{ce} + \lambda_{1} L_{mgn} + \lambda_2 L_{reg}$, where $L_{ce}$ is the standard cross-entropy loss, $L_{reg}$ is the L2 regularization term and $L_{mgn}$ is the inter-task separation loss, $\lambda_{1}$ and $\lambda_{2}$ are the importance weights for each loss.

% \textbf{Inference}
% Upon finishing of training, FSCN can be used to refine all proposals from the last stage of Faster-RCNN. Specially, we propose to fuse their classification score via:
% \begin{equation}
% \begin{aligned}
% s=s_{detector} \odot s_{corrector}
% \end{aligned}
% \end{equation}
% \normalsize
% where $s_{detector}$ and $s_{corrector}$ denote the softmax classification score of the base detector and FSCN, respectively.

\section{Experiments}

\subsection{ Dataset Settings}
Follow the common practice of few-shot detection, we combine the 80K images training set and 35K images trainval set of MS-COCO 2014 to train our model, and report evaluation results based on the 5K minival set. We then split the overall 80 categories into two groups, 20 classes which are shared with Pascal VOC are used as novel classes, and the other 60 classes are used as base classes. To evaluate the robustness of our method against forgetting, we design a challenging incremental learning setting, where the 20 novel classes are added one by one through 20 incremental steps. In practice, it is worth to note that new-class images may also contain instances belonging to old classes. However, we do not provide any annotations for old-class instances during model adaption. That is, upon the arrival of each novel class, the model can only access to k = 10 bounding-box annotation of the current class. We then evaluate our model over all classes and report detection performance of base and novel classes separately.

For another cross-domain evaluation from MS-COCO to VOC, model is first pre-trained on the 60 base classes from MS-COCO. Then Pascal VOC 07+12 training set is employed as novel set for model adaption with 20 incremental steps. Next, we test our method to detect the 20 novel classes based on VOC2007 test set. Different from the first experiment that focuses on evaluating cross-category model generalization, this setup is further to appraise the cross-domain generalization ability.

\subsection{Baseline settings}
\textbf{Meta-learning approaches}
We first compare our approach to multiple few-shot detection approaches. The first two baselines, "Feature-Reweight~\cite{Kang_2019_YOLOLS}" and "ONCE ~\cite{Once}" belong to meta-learning approaches that predict detections conditioned on a set of support examples. For instance, "Feature-Reweight" propose to reweight the deep layers of a CNN backbone with channel-wise attentive vectors, so as to amplify those informative features for detecting novel objects in the query set. "ONCE" propose a kernel-weight generator that extracts meta-information from support samples to predict convolution-kernel weights for detecting novel classes. The major drawback of these meta-learning approaches is that, during multi-step adaption, the meta-learner must be fine-tuned with sufficient training categories for generalization. On the one hand, this requires recalling historical training samples which may cause both storage and privacy issues. On the other hand, as their computational complexity is proportional to the number of categories, they will become rather slow or even unavailable when tackling continually increasing data sets.

\textbf{Transfer-learning approaches}
We then compare our approach with two transfer-learning based approaches: TFA and IMTFA~\cite{IMTFA}. TFA alleviates catastrophic forgetting on both base and novel classes by its two-stage finetuning approach, in which only the last layer of box classification and regression is finetuned with cosine similarity while freezing backbone, RPN and ROI feature extractor. IMTFA improves TFA by setting new classification weights through weight imprinting rather than randomly initialization. However, the fixed feature extractor is biased towards features that are discriminative and salient to base classes, thus freezing it without exposure to novel classes can be suboptimal.

\textbf{Conventional incremental learning approaches}
Compared with the general incremental learning method, our method is more suitable for the sequential adaptation with small training sets. To prove this point, we compare
it with the incremental learning framework Specifically, we use the naïve fine-tuning method as baseline, denotes as FRCN-ft, where we unfreeze the ROI feature extractor to fine-tune with novel classes. Then we embed different incremental learning framework into the baseline method FRCN-ft,e.g, IMM~\cite{IMM} and ILOD~\cite{ILOD}.
\begin{table}
    \caption{Per-Class Incremental Results on MS-COCO.}
    \label{table:Per-Class COCO}
\centering
\setlength{\tabcolsep}{1.5mm}{
\begin{tabular}{clcc}
\toprule
\toprule
\multicolumn{2}{l}{\multirow{2}{*}{}}                     & \multicolumn{2}{c}{10-Shot}   \\

\multicolumn{2}{l}{}                                          & Novel AP     & Base AP        \\
\hline
\multirow{10}{*}{Method} & Feature-Reweight                        & 1.5          & 3.7            \\
                         & ONCE                                 & 1.2          & 17.9           \\
                         & FRCN-ft                                    & 3.2          & 17.9           \\
                         & TFA                                   & 5.6          & 27.8           \\
                         & IMTFA                                   & 6.1          & 28.0           \\
                         & IMM                                   & 5.3          & 21.5           \\
                         & ILOD                                 & 4.9          & 20.7           \\
                         & SMM                                   & 7.4          & 27.6           \\
                         & SMM+CR                                   & 7.9          & 27.3           \\
                         & SMM+CR+Inter-Sep                         & 8.4          & 26.8           \\
                         & DBF+SMM+CR+Inter-Sep  & \textbf{8.5} & \textbf{28.1} \\
                         \bottomrule
\end{tabular}}
\end{table}

\begin{table}
    \caption{Per-Group Incremental Results on MS-COCO.}
    \label{table:Per-Group COCO}
\centering
\setlength{\tabcolsep}{1.5mm}{
\begin{tabular}{clcc}
\toprule
\toprule
\multicolumn{2}{l}{\multirow{2}{*}{}}                      & \multicolumn{2}{c}{10-Shot}   \\

\multicolumn{2}{l}{}                                           & Novel AP     & Base AP        \\
\hline
\multirow{10}{*}{Method} & Feature-Reweight                       & -          & -           \\
                         & ONCE                                   & -          & -           \\
                         & FRCN-ft                                    & 4.3          & 21.6           \\
                         & TFA                                    & 6.0          & 28.5           \\
                         & IMTFA                                  & 6.4          & \textbf{28.7}           \\
                         & IMM                                   & 7.1          & 24.1           \\
                         & ILOD                                   & 6.6          & 23.6           \\
                         & SMM                                    & 8.2          & 27.9           \\
                         & SMM+CR                                    & 8.5          & 27.7           \\
                         & SMM+CR+Inter-Sep                         & 9.1          & 27.3           \\
                         & DBF+SMM+CR+Inter-Sep &  \textbf{9.1} & 28.5 \\
                         \bottomrule
\end{tabular}}
\end{table}

\begin{table}
    \caption{Comparison with baselines from MS-COCO to Pascal VOC.}
\centering
\setlength{\tabcolsep}{1mm}{

 \begin{tabular}{*{22}{c}}
    \toprule
    \toprule
        & \multirow{1}{*}{{Methods}}

    &        & 5-Shot & 10-Shot         \\
    \midrule

  %& MAML  &   &   0.6  &   1.0  \\
  & ONCE   &   & 2.4 & 2.6     \\
  & TFA   &   & 6.7 & 10.6     \\
  & IMTFA  &   & 7.3 & 12.4     \\
   & ours &   &  \textbf{12.5}  & \textbf{23.7}       \\
    \bottomrule
    \end{tabular}}
\label{table:COCO to VOC}
\end{table}
\subsection{Results on MS-COCO}
Now, we compare the proposed DBF framework with the previous state-of-the-art methods. A theoretical upper bound for iFSOD is the results of non-incremental training based on the same data, where model is joint training with all classes without considering incomplete annotations, while the lower bounds are obtained from continually fine-tuning a naïve Faster-RCNN model with multiple small data sets, we denote it as FRCN-ft. Except for the upper-bound method, the phenomenon of incomplete annotations is set to be appeared in all other baselines, thus for a fair comparison, the proposed semi-supervised object mining (SSOM) method is embedded into all baselines.We compare the methods under the following two scenarios, where novel class may come one by one or as a group.

\textbf{Per-Class Incremental}
To compare our method with the other baselines towards the capability of preserving long-term memory, we first pre-train on the 60 base classes and then adapt to the remaining 20 novel classes step by step, where each step only contains one novel classes. Regarding the results, we have several observations. 1) The accuracy of the existing meta-learning approaches is still far away from satisfaction. Moreover, fine-tuning meta learners with reduced number of class is infeasible since the episodic learning scheme always requires category diversity, thus making them not feasible for real-world iFSOD scenarios. 2) Compared with the baseline TFA, unfreezing RoI feature extractor (FRCN-ft) gets even worse results, which indicates that naively fine-tuning more parameters with limited data may aggravate overfitting and then cause catastrophic forgetting. (3) IMM can relieve catastrophic forgetting to some extent. However, such improvement is still marginal since IMM only merges models rather than restricting the overall trajectory of parameter updating. 4) The proposed SMM approach outperforms the original IMM approach significantly under 10-shots, which indicates that restricting the overall parameter displacement is crucial for long-term memory. 5)Thanks to the effort of SMM that always maintains a flat local minimum during continually fine-tuning, the  degree of feature drift of embedding-space can be effectively reduced to minimum. Therefore, the last-layer feature vectors(SMM+CR) can be used as effective class representatives for representing old-class distributions, which brings 0.4 mAP improvements, this also relieves the requirements of storing raw images. 6)The proposed inter-task class(SMM+CR+ Inter-Sep) separation loss consistently performs better than cross-entropy only, which indicates that a more compact intra-class representation is formed by learning a large margin between old and new classes.  Thanks to the non-forgetting property, our proposed methods gain +2.4 AP for 10-shot above current SOTA, which is more significant than the gaps between any previous advancements.

\textbf{Per-Group Incremental}
We then test our method in a group incremental scenario where the 20 novel classes are divided into four groups with five classes in each group. The incremental fine-tuning is conducted by adding one group of classes at each time until all 20 classes are learned. The results are shown in Table, where our method significantly outperforms the other baselines at each group, confirming that our method is also superior in the group-wise incremental detection setting.

\subsubsection{Results on MS-COCO to Pascal VOC}We then evaluated our method in a cross-domain setting, where the data used for model pre-training and incremental fine-tuning are from different domains. Specially, we first train a model on the 60 base classes from MS-COCO, then we fine-tune it step by step on the 20 novel classes from Pascal VOC, and evaluate it on VOC2007 test set. The results in Table~\ref{table:COCO to VOC} confirm the generalization advantages of our method when transfers to a novel domain different from the training one.

%\begin{table}[h]
 %   \caption{Results of different incremental steps on MS-COCO.}
 %   \label{table:incremental COCO}
%\centering
%\setlength{\tabcolsep}{1.5mm}{
%\begin{tabular}{clcccc}
%\toprule
%\toprule
%\multicolumn{2}{l}{\multirow{2}{*}{}}                   & \multicolumn{2}{c}{20-step}   & \multicolumn{2}{c}{4-step}   \\
%\cline{3-6}
%\multicolumn{2}{l}{}                                    & Novel AP     & Base AP       & Novel AP     & Base AP        \\
%\hline
%\multirow{10}{*}{}

 %                        & TFA-fc (baseline)                       & 5.8          & 28.6          & 5.5          & 28.6           \\

  %                       & DBF + mean-IMM (baseline)    & 4.6          & 28.5          & 5.9          & 28.6           \\
   %                      & DBF + SMM (ours)             & 6.9          & 28.6          & 6.8          & \textbf{28.6}           \\
%                         & DBF + SMM + Inter-Sep (ours) & \textbf{7.9} & \textbf{28.6} & \textbf{8.5} & 28.5 \\
 %                        \bottomrule
%\end{tabular}}
%\end{table}

\section{Conclusion}
We propose a generic learning scheme for CI-FSOD. First, the Double Branch Framework preserves base-class feature distribution for performance retention. Second, the Stable Moment Matching method solves the issue of catastrophic forgetting under the setting of few-shot, which provides a better trade-off between stability and plasticity. Third, the proposed Inter-Task Class Separation promotes a large separation between old and new classes. The effectiveness is validated by extensive experiments, where our approach yields the state-of-the-art performance and outperform the previous algorithm with a significant margin.

\bibliographystyle{ieeetr}
\bibliography{LearningClassificationRefinementForLowShotObjectDetection}

\begin{thebibliography}{10}

\bibitem{NIPS2015_FasterRCNN}
S.~Ren, K.~He, R.~Girshick, and J.~Sun, ``Faster {R-CNN}: Towards real-time
  object detection with region proposal networks,'' in {\em Advances in Neural
  Information Processing Systems 28}, pp.~91--99, 2015.

\bibitem{Bochkovskiy_2020_Yolo4}
A.~Bochkovskiy, C.-Y. Wang, and H.-Y.~M. Liao, ``{YOLOv4}: Optimal speed and
  accuracy of object detection,'' 2020.

\bibitem{cascade_rcnn}
Z.~Cai and N.~Vasconcelos, ``Cascade r-cnn: Delving into high quality object
  detection,'' {\em 2018 IEEE/CVF Conference on Computer Vision and Pattern
  Recognition}, 2018.

\bibitem{Wang2020}
X.~{Wang}, T.~E. {Huang}, T.~{Darrell}, J.~E. {Gonzalez}, and F.~{Yu},
  ``{Frustratingly Simple Few-Shot Object Detection},'' {\em arXiv e-prints},
  2020.

\bibitem{Yan2019MetaRCNN}
X.~Yan, Z.~Chen, A.~Xu, X.~Wang, X.~Liang, and L.~Lin, ``{Meta R-CNN}: Towards
  general solver for instance-level low-shot learning,'' pp.~9576--9585, 10
  2019.

\bibitem{Once}
J.-M. Perez-Rua, X.~Zhu, T.~Hospedales, and T.~Xiang, ``Incremental few-shot
  object detection,'' 2020.

\bibitem{BBN}
B.~Zhou, Q.~Cui, X.-S. Wei, and Z.-M. Chen, ``Bbn: Bilateral-branch network
  with cumulative learning for long-tailed visual recognition,'' in {\em
  Proceedings of the IEEE/CVF Conference on Computer Vision and Pattern
  Recognition}, pp.~9719--9728, 2020.

\bibitem{understanding_IL}
S.~I. Mirzadeh, M.~Farajtabar, R.~Pascanu, and H.~Ghasemzadeh, ``Understanding
  the role of training regimes in continual learning,'' {\em arXiv preprint
  arXiv:2006.06958}, 2020.

\bibitem{CPR}
S.~Cha, H.~Hsu, T.~Hwang, F.~P. Calmon, and T.~Moon, ``Cpr:
  Classifier-projection regularization for continual learning,'' {\em arXiv
  preprint arXiv:2006.07326}, 2020.

\bibitem{IMM}
S.-W. Lee, J.-H. Kim, J.~Jun, J.-W. Ha, and B.-T. Zhang, ``Overcoming
  catastrophic forgetting by incremental moment matching,'' {\em arXiv preprint
  arXiv:1703.08475}, 2017.

\bibitem{Chelsea_2017_MAML}
C.~Finn, P.~Abbeel, and S.~Levine, ``Model-agnostic meta-learning for fast
  adaptation of deep networks,'' in {\em ICML}, 2017.

\bibitem{Snell_2017_Prototypical}
J.~Snell, K.~Swersky, and R.~Zemel, ``Prototypical networks for few-shot
  learning,'' in {\em Advances in Neural Information Processing Systems 30},
  2017.

\bibitem{IDQ}
K.~Chen and C.-G. Lee, ``Incremental few-shot learning via vector quantization
  in deep embedded space,'' in {\em International Conference on Learning
  Representations}, 2021.

\bibitem{EWC}
J.~Kirkpatrick, R.~Pascanu, N.~Rabinowitz, J.~Veness, G.~Desjardins, A.~A.
  Rusu, K.~Milan, J.~Quan, T.~Ramalho, A.~Grabska-Barwinska, {\em et~al.},
  ``Overcoming catastrophic forgetting in neural networks,'' {\em Proceedings
  of the national academy of sciences}, vol.~114, no.~13, pp.~3521--3526, 2017.

\bibitem{Hao2018LSTD}
H.~Chen, Y.~Wang, G.~Wang, and Y.~Qiao, ``{LSTD}: A low-shot transfer detector
  for object detection,'' in {\em AAAI}, 2018.

\bibitem{Kang_2019_YOLOLS}
B.~Kang, Z.~Liu, X.~Wang, F.~Yu, J.~Feng, and T.~Darrell, ``Few-shot object
  detection via feature reweighting.,'' in {\em ICCV}, 2019.

\bibitem{linearmultiple}
S.~I. Mirzadeh, M.~Farajtabar, D.~Gorur, R.~Pascanu, and H.~Ghasemzadeh,
  ``Linear mode connectivity in multitask and continual learning,'' {\em arXiv
  preprint arXiv:2010.04495}, 2020.

\bibitem{linearconnectivity}
J.~Frankle, G.~K. Dziugaite, D.~Roy, and M.~Carbin, ``Linear mode connectivity
  and the lottery ticket hypothesis,'' in {\em International Conference on
  Machine Learning}, pp.~3259--3269, PMLR, 2020.

\bibitem{keskar2016large}
N.~S. Keskar, D.~Mudigere, J.~Nocedal, M.~Smelyanskiy, and P.~T.~P. Tang, ``On
  large-batch training for deep learning: Generalization gap and sharp
  minima,'' {\em arXiv preprint arXiv:1609.04836}, 2016.

\bibitem{Hang2018}
H.~Qi, M.~Brown, and D.~G. Lowe, ``Low-shot learning with imprinted weights,''
  {\em 2018 IEEE/CVF Conference on Computer Vision and Pattern Recognition},
  pp.~5822--5830, 2018.

\bibitem{semantic_drift}
L.~Yu, B.~Twardowski, X.~Liu, L.~Herranz, K.~Wang, Y.~Cheng, S.~Jui, and
  J.~v.~d. Weijer, ``Semantic drift compensation for class-incremental
  learning,'' in {\em Proceedings of the IEEE/CVF Conference on Computer Vision
  and Pattern Recognition}, pp.~6982--6991, 2020.

\bibitem{cosface}
H.~Wang, Y.~Wang, Z.~Zhou, X.~Ji, D.~Gong, J.~Zhou, Z.~Li, and W.~Liu,
  ``Cosface: Large margin cosine loss for deep face recognition,'' 2018.

\bibitem{IMTFA}
D.~A. Ganea, B.~Boom, and R.~Poppe, ``Incremental few-shot instance
  segmentation,'' in {\em Proceedings of the IEEE/CVF Conference on Computer
  Vision and Pattern Recognition}, pp.~1185--1194, 2021.

\bibitem{ILOD}
K.~Shmelkov, C.~Schmid, and K.~Alahari, ``Incremental learning of object
  detectors without catastrophic forgetting,'' in {\em Proceedings of the IEEE
  international conference on computer vision}, pp.~3400--3409, 2017.

\end{thebibliography}

\end{document}